\documentclass[10pt,twocolumn,letterpaper]{article}

\usepackage{iccv}
\usepackage{times}
\usepackage{epsfig}
\usepackage{graphicx}
\usepackage{amsmath}
\usepackage{amssymb}

\usepackage{booktabs}
\usepackage{multirow}
\usepackage{bm}
\usepackage{tikz}
\usepackage{comment}
\usepackage{color}
\usepackage{bbding}
\usepackage{gensymb}
\usepackage{colortbl}
\usepackage{algorithmicx,algorithm}
\usepackage[noend]{algpseudocode}
\usepackage[accsupp]{axessibility} 

\usepackage[breaklinks=true,colorlinks,bookmarks=false]{hyperref}

\iccvfinalcopy 

\def\iccvPaperID{5562} 

\ificcvfinal\fi

\begin{document}

\title{VI-Net: Boosting Category-level 6D Object Pose Estimation via Learning Decoupled Rotations on the Spherical Representations}

\author{
   Jiehong Lin$^{1,2}$ \and Zewei Wei$^1$ \and Yabin Zhang$^3$ \and Kui Jia$^1$\thanks{Correspondence to Kui Jia $<$kuijia@gmail.com$>$} \and
    $^1$South China University of Technology  \\ $^2$DexForce Technology Co. Ltd. \quad  \quad  $^3$Hong Kong Polytechnic University
   \\ {\tt\small \url{https://github.com/JiehongLin/VI-Net}}
}

\maketitle
\ificcvfinal\thispagestyle{empty}\fi

\begin{abstract}
   Rotation estimation of high precision from an RGB-D object observation is a huge challenge in 6D object pose estimation, due to the difficulty of learning in the non-linear space of $SO(3)$.
   In this paper, we propose a novel rotation estimation network, termed as \textbf{VI-Net}, to make the task easier by decoupling the rotation as the combination of a viewpoint rotation and an in-plane rotation. More specifically, VI-Net bases the feature learning on the sphere with two individual branches for the estimates of two factorized rotations, where a V-Branch is employed to learn the viewpoint rotation via binary classification on the spherical signals, while another I-Branch is used to estimate the in-plane rotation by transforming the signals to view from the zenith direction. 
   To process the spherical signals, a Spherical Feature Pyramid Network is constructed based on a novel design of \textit{SPAtial Spherical Convolution} (SPA-SConv), which settles the boundary problem of spherical signals via feature padding and realizes \emph{viewpoint-equivariant} feature extraction by symmetric convolutional operations.
   We apply the proposed VI-Net to the challenging task of category-level 6D object pose estimation for predicting the poses of unknown objects without available CAD models; experiments on the benchmarking datasets confirm the efficacy of our method, which outperforms the existing ones with a large margin in the regime of high precision. 
\end{abstract}

\section{Introduction}
\label{sec:intro}

The task of 6D object pose estimation \cite{xiang2017posecnn, wang2019densefusion, he2021ffb6d, li2022dcl, NOCS} from an RGB-D object observation is to learn the transformation from the canonical object system to the camera system, represented by a 3D rotation $\bm{R}\in SO(3)$ and a 3D translation $\bm{t} \in \mathbb{R}^3$. It is demanded in many real-world applications, such as robotic grasping \cite{wu2020grasp,mousavian20196}, augmented reality \cite{azuma1997survey}, and autonomous driving \cite{levinson2011towards,wang2019frustum,chen2017multi,deng2022vista}.

For the 6D object pose, translation is easier to be estimated, \eg, initialized as the centroid of the object points, while learning rotation in the whole space of $SO(3)$ is more challenging due to the non-linearity of $SO(3)$. Things become more complicated when estimating rotations of unknown objects without available CAD models. 
Taking the task of category-level 6D object pose estimation as an example, 
one typical way  \cite{NOCS, SPD, CRNet, SGPA} is to learn the correspondence between the camera system and the canonical one to solve the poses via Umeyama algorithm\cite{Umeyama}, with the surrogate objectives of canonical coordinates rather than the true ones of object poses.
Another strategy is to learn in the $SO(3)$ space with rotation-aware features extracted from specially designed encoders, \eg, 3D Graph Convolutional (3DGC) autoencoder \cite{FSNet}, Spherical CNN (in the spectral domain) \cite{esteves2018learning,lin2021dualposenet}, and 3D Steerable CNN \cite{SS-Conv}, whose results, however, are less satisfactory than those based on correspondence learning.

\begin{figure}[t]
  \centering
   \includegraphics[width=0.98\linewidth]{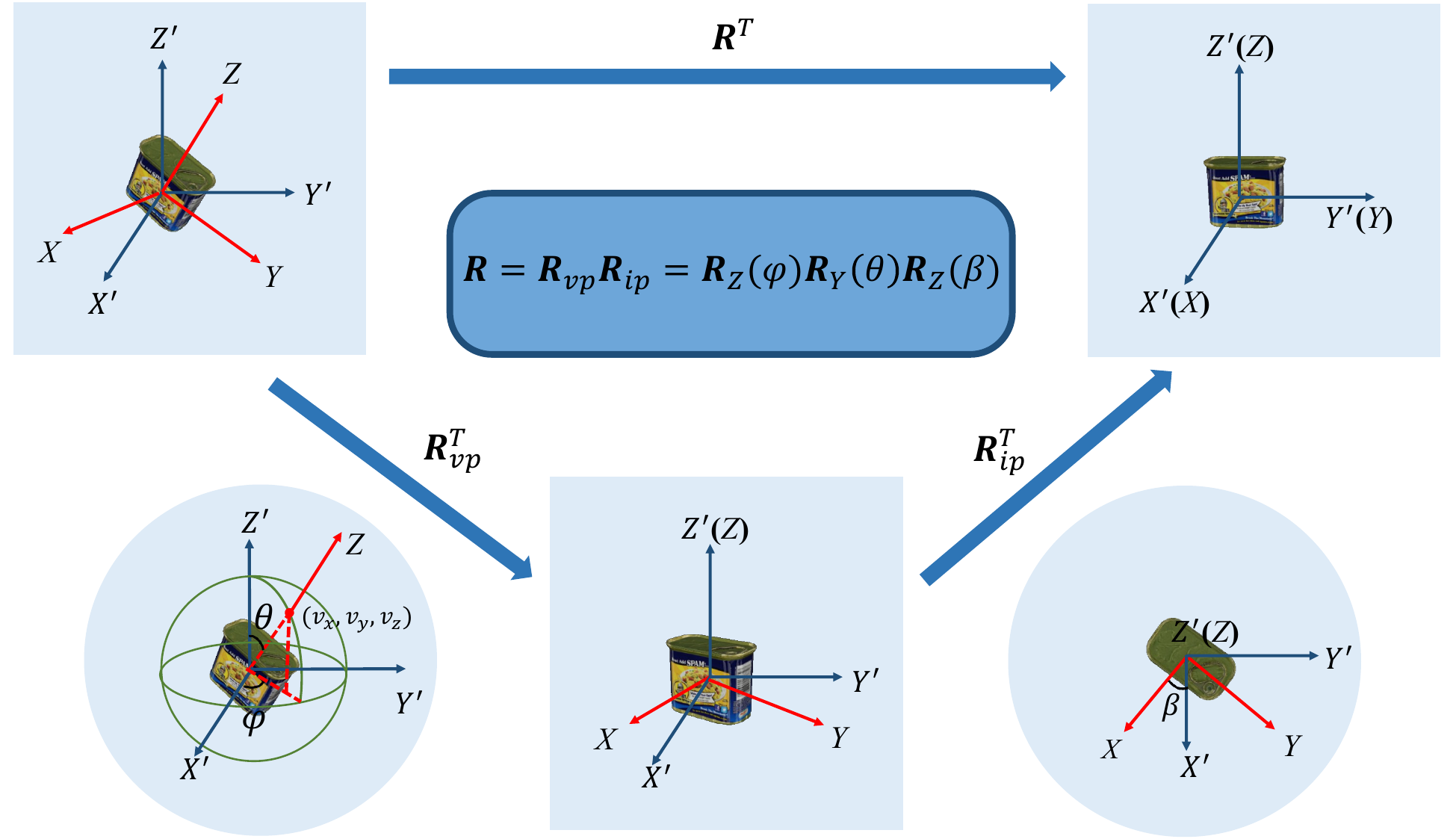}
   \caption{An illustration of the factorization of rotation $\bm{R}$ into a viewpoint (out-of-plane) rotation $\bm{R}_{vp}$ and an in-plane rotation $\bm{R}_{ip}$ (around $Z$-axis). Notations are explained in Sec. \ref{sec:correlation_on_sphere_and_rotation_decomposition}.}
   \vspace{-0.5cm}
   \label{fig:intro}
\end{figure}

Recently, OVE6D \cite{cai2022ove6d} narrows the learning space of rotation $\bm{R}$ by factorizing it, as shown in Fig. \ref{fig:intro}, into a viewpoint rotation $\bm{R}_{vp}$ and an in-plane rotation $\bm{R}_{ip}$ around the canonical zenith direction (the positive direction of $Z$-axis in Fig. \ref{fig:intro}), and renders images of object CAD models with various of discrete rotations to construct the codebook for viewpoint rotation retrieval. However, object CAD models are often not available in real-world applications for rendering, \eg, on the focused task of category-level 6D object pose estimation. In this paper, we correlate the factorized rotations to the sphere, and propose a new method of \textbf{VI-Net}, which employs two carefully designed heads of \textbf{V-Branch} and \textbf{I-Branch} to estimate $\bm{R}_{vp}$ and $\bm{R}_{ip}$, respectively, based on spherical representations with no use of object CAD models; Fig. \ref{fig:vi-net} gives an illustration of VI-Net.

Specifically, given an object observation (centered at the origin with known translation), VI-Net processes its point-wise attributes as the signals assigned on the sphere, represented as a 2D spherical map, and constructs a Spherical Feature Pyramid Network (FPN) to extract high-level spherical features along hierarchy; on top of the Spherical FPN, two individual heads of V-Branch and I-Branch are employed to learn the viewpoint rotation $\bm{R}_{vp}$ and the in-plane rotation $\bm{R}_{ip}$, respectively. 
For V-Branch, $\bm{R}_{vp}$ is generated by learning the intersection point of the canonical zenith direction with the unit sphere, which is recognized via binary classification on the high-level spherical feature map.
For I-Branch, VI-Net rotates the object through the transformation of its spherical feature map with $\bm{R}_{vp}$, and could thus estimate $\bm{R}_{ip}$ from the perspective of the canonical zenith direction. Finally, the rotation $\bm{R}$ is given as the multiplication of $\bm{R}_{vp}$ and $\bm{R}_{ip}$.

Another problem is how to build the Spherical FPN. For convenience, the spherical signals are processed in the form of representations with regular 2D spatial sizes, which yet results in the boundary problem. 
We thus propose in this paper a novel design of \textit{SPAtial Spherical Convolution}, termed as \textbf{SPA-SConv}, which settles the above problem via simple feature padding, and furthermore, extracts viewpoint-equivariant features through symmetric convolutional operations to support the feature transformation in I-Branch.
Our SPA-SConv could be flexibly adapted to the existing convolutional architectures by replacing the convolutions, and we thus construct the spherical version of FPN \cite{lin2017feature}.

To confirm the efficacy of our VI-Net on rotation estimation, we apply it to the challenging task of category-level 6D object pose estimation. Experiments are conducted on the benchmarking REAL275 dataset \cite{NOCS}, which show that our method outperforms the existing ones with a large margin in the regime of high precision, \eg, an improvement of $4.0\%$ on the metric of $5\degree 2$cm over the state-of-the-art method of DPDN \cite{DPDN}. Ablation studies also confirm the efficacy of our novel designs.

\section{Related Work}
\label{sec:related}

\subsection{Category-level 6D Object Pose Estimation}

The task of category-level 6D object pose estimation is introduced in \cite{NOCS} to estimate the 6D pose and size of unseen objects within certain categories, and could be categorized into two groups, including those based on correspondence learning and those based on direct regression.

\noindent \textbf{Methods Based on Correspondence Learning} This group of methods first learn the points in the canonical space to align with the observations, and then obtain the object poses from the correspondence via solving of Umeyama algorithm \cite{Umeyama}. NOCS \cite{NOCS} regresses the canonical point coordinates directly from an RGB image via MaskRCNN \cite{MasRCNN}, while SPD \cite{SPD} learns to deform the categorical shape priors to improve the qualities of canonical points. The latter realizes better correspondence by taking the advantage of shape priors, and attracts the followers to embark on this path. For example, CR-Net \cite{CRNet} models the relationship between the observations and the shape priors, and iteratively refines the shape deformations, while SGPA \cite{SGPA} proposes a structured-guided prior adaptation scheme via transformers.

\noindent \textbf{Methods Based on Direct Regression} Most methods of this group directly regress the object poses from the rotation-aware features via specially designed network architectures. DualPoseNet \cite{lin2021dualposenet} employs the $SO(3)$-equivariant spherical convolutions for the extraction of the global pose-sensitive features, on top of which two pose decoders with different working mechanisms are designed for complementary supervision. 3D Steerable Convolutions are proposed in \cite{SS-Conv} for $SE(3)$-equivariant feature learning, and successfully applied to this task. Both FS-Net \cite{FSNet} and GPV-Pose \cite{GPV} construct their encoders based on 3D Graph Convolutions (3DGC), from which two perpendicular vectors of rotation are directly regressed; we note that they also decouple the rotation to reduce the learning space, but do not well exploit the properties of the perpendicular vectors for easier learning.
Recently, DPDN \cite{DPDN} borrows the idea of correspondence learning to deform the features of shape priors, and builds correspondence in the feature space for direct estimates of object poses.

\vspace{-0.1cm}
\subsection{Convolutions on Spherical Representations}
\vspace{-0.1cm}
\label{subsec:related_ssconv}
The early works \cite{su2017learning, boomsma2017spherical, zhao2018distortion} of convolutions, defined on spherical representations, consider less about the properties of equivariance, which, yet, are important for rotation estimation.
Then Esteves et al. propose a design of spherical convolution \cite{esteves2018learning} implemented in the spectral domain to learn features with strict $SO(3)$-equivariant constraints, which is used in \cite{lin2021dualposenet} for pose estimation and most related to our SPA-SConv; to distinguish two kinds of convolutions, we term their design as \textbf{SPE-SConv}. In our VI-Net, SO(3)-equivariance of SPE-SConv makes the individual feature points not discriminative to support the binary classification for generating the viewpoint rotation, even resulting in gradient exploding; our SPA-SConv relaxes the constraints while preserving viewpoint-equivariance, which could jointly support the binary classification in V-Branch and the feature transformation in I-Branch.

\section{Correlation on the Sphere and Rotation Decomposition}
\label{sec:correlation_on_sphere_and_rotation_decomposition}

A rotation $\bm{R} \in SO(3)$ could be defined as the transformation between two Euclidean coordinate systems with the same origins, \eg, from the canonical $XYZ$ system to the observed $X^{\prime}Y^{\prime}Z^{\prime}$ one in Fig. \ref{fig:intro}. One can determine it with the positive direction of $Z$-axis in $X^{\prime}Y^{\prime}Z^{\prime}$ system (the canonical zenith direction of the $XYZ$ system) and the in-plane rotation angle $\beta \in [0, 2\pi]$ around the $Z$-axis, which is exactly the decomposition of $\bm{R}$ as multiplication of an out-of-plane (viewpoint) rotation $\bm{R}_{vp}$ and a 2D in-plane rotation $\bm{R}_{ip}$ \cite{cai2022ove6d}:
\begin{equation}
    \bm{R} = \bm{R}_{vp}  \bm{R}_{ip},
\end{equation}
with
\begin{equation}
    \bm{R}_{ip} = \bm{R}_Z(\beta) = \begin{bmatrix}
        \cos \beta  & -\sin \beta & 0\\
        \sin \beta & \cos \beta & 0\\
        0 & 0 & 1 
       \end{bmatrix}.
       \label{eqn:r_ip}
\end{equation}

In the $X^{\prime}Y^{\prime}Z^{\prime}$ system, let $\bm{v}$ be the final column of the $3\times3$ rotation matrix $\bm{R}$, which is a unit vector ending at the point $(v_x, v_y, v_z)$ on a unit sphere $S^2$; the direction of $\bm{v}$ is aligned with the zenith direction of $XYZ$ system. Transforming the $X^{\prime}Y^{\prime}Z^{\prime}$ system to spherical coordinate system, we have the spherical coordinate $(r, \varphi, \theta)$ of the point $(v_x, v_y, v_z)$ as follows:
\begin{equation}
    \left\{\begin{aligned}
    &r = 1    \\
    &\varphi = \arctan(v_x/v_z) \\
    &\theta = \arccos(v_y/r)  
       \end{aligned}\right.,
\end{equation}
where $\varphi \in [0, 2\pi] $ and $\theta \in [0, \pi]$ are the azimuthal and inclination angles, respectively. Then we can compute $\bm{R}_{vp}$ as follows:
\begin{align}
    \bm{R}_{vp} = &  \bm{R}_Z(\varphi) \bm{R}_Y(\theta) \notag \\
    = & \begin{bmatrix}
        \cos \varphi  & -\sin \varphi & 0\\
        \sin \varphi & \cos \varphi & 0\\
        0 & 0 & 1 
       \end{bmatrix} \times \begin{bmatrix}
        \cos \theta & 0 & \sin \theta\\
        0 & 1 & 0\\
        -\sin \theta & 0 & \cos \theta
       \end{bmatrix}.
       \label{eqn:r_vp}      
\end{align}
Combining (\ref{eqn:r_vp}) and (\ref{eqn:r_ip}), we can find that the rotation decomposition $\bm{R} = \bm{R}_{vp}  \bm{R}_{ip} $ is consistent with the parameterization of $\bm{R}$ with $ZYZ$-Euler angles $\varphi$, $\theta$ and $\beta$.

Motivated by the correlation on the sphere and rotation decomposition, we thus propose in this paper to base the feature learning on the sphere and decouple the estimation of $\bm{R}$, with no use of object CAD models, into two parts:
\begin{itemize}
    \item Searching on the sphere for the point $(v_x, v_y, v_z)$ to obtain the angles $\varphi$ and $\theta$, which jointly give the viewpoint rotation $\bm{R}_{vp}$;
    \item Aligning $Z^{\prime}$-axis with $Z$-axis by $\bm{R}_{vp}$ to view from the canonical zenith direction with fewer learning patterns, and then regressing the in-plane rotation $\bm{R}_{ip}$.
\end{itemize}

\section{VI-Net for Rotation Estimation}
\label{sec:method}

According to the discussion in Sec. \ref{sec:correlation_on_sphere_and_rotation_decomposition}, we propose VI-Net for the estimation of rotation $\bm{R}$, as shown in Fig. \ref{fig:vi-net}. Specifically, given a point set $\mathcal{P}$ of a target object, spherical input signals (Sec. \ref{subsec:data_processing}) are generated for the extraction of high-level viewpoint-equivariant spherical features $\mathcal{S}$ via a Spherical Feature Pyramid Network (Sec. \ref{subsec:spherical_feature_prymarid_network});
on top of $\mathcal{S}$, VI-Net decouples the rotation estimation into two branches: the V-Branch learns the viewpoint rotation $\bm{R}_{vp}$ via binary classification on each sampled grid of the sphere (Sec. \ref{subsec:V_branch}), while the I-Branch estimates the in-plane rotation $\bm{R}_{ip}$ by transforming $\mathcal{S}$ to view the object from the canonical zenith direction (Sec. \ref{subsec:I_branch}); finally, the rotation $R$ is given out as the multiplication of  $\bm{R}_{vp}$ and  $\bm{R}_{ip}$. We also provide the training objective in Sec. \ref{subsec:objective}.

\subsection{Conversion as Spherical Representations}
\label{subsec:data_processing}

Given a point set $\mathcal{P} \in \mathbb{R}^{N\times3}$ with its point-wise attributes $\mathcal{F} \in \mathbb{R}^{N\times C_0}$ (\eg, radial distance, RGB values, surface normal, \etc), we firstly follow \cite{esteves2018learning, lin2021dualposenet} to generate the feature map $\mathcal{S}_0 \in \mathbb{R}^{C_0\times H_0 \times W_0}$ defined on the sphere, where $N$, $H_0 \times W_0$ and $C_0$ are the point number, the spherical sampling resolution, and attribute dimension (\eg, $C_0=1$ for radial distance and $C_0=3$ for RGB values or surface normal), respectively.

More specifically, we firstly divide, in the spherical coordinate system, $W_0$ and $H_0$ bins uniformly along the azimuthal and inclination axes, respectively, resulting in $H_0 \times W_0$ regions in the space (\cf Fig. \ref{fig:spa-sconv}). Within the region indexed by $(h, w)$, we search for the point with the largest radial distance, denoted as $\bm{p}^{max}_{h,w}$, and set $\mathcal{S}_0(h,w) = \bm{f}_{h,w}^{max} \in \mathcal{F}$, which corresponds to $\bm{p}^{max}_{h,w}$; if there is no point in the region, we set $\mathcal{S}_0(h,w) = \bm{0}$.

\textbf{\begin{figure*}[t]
  \centering
  \includegraphics[width=\linewidth]{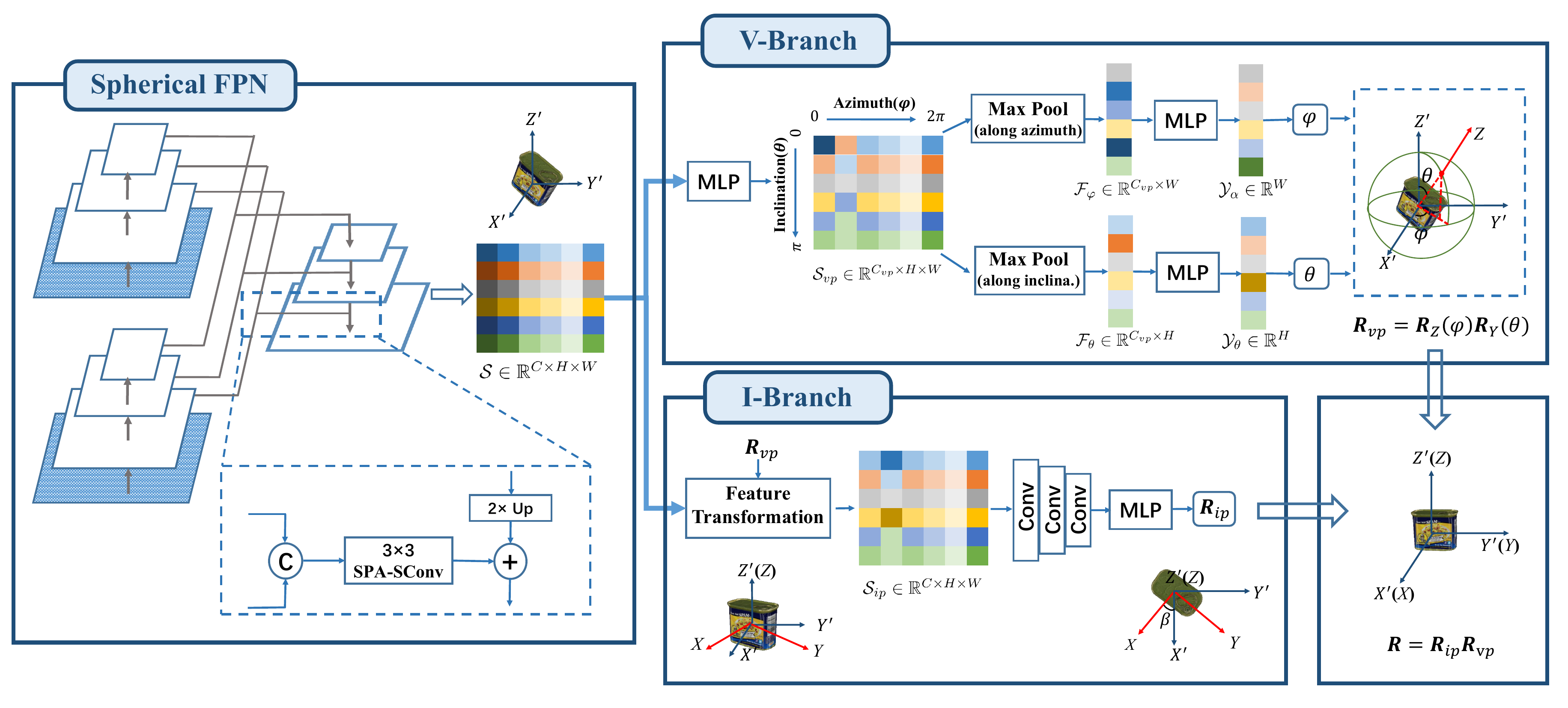}
   \caption{An illustration of VI-Net for rotation estimation. We firstly construct a Spherical Feature Pyramid Network based on spatial spherical convolutions (SPA-SConv) to exact the high-level spherical feature map $\mathcal{S}$. On top of $\mathcal{S}$, a \textbf{V-Branch} is employed to search the canonical zenith direction on the sphere via binary classification for the generation of the viewpoint rotation $\bm{R}_{vp}$, while another \textbf{I-Branch} is used to estimate the in-plane rotation $\bm{R}_{ip}$ by transforming $\mathcal{S}$ to view the object from the canonical zenith direction. Finally we have $\bm{R}=\bm{R}_{vp}\bm{R}_{ip}$. Best view in the electronic version.}
   \label{fig:vi-net}
\end{figure*}}

\subsection{Spherical Feature Pyramid Network}
\label{subsec:spherical_feature_prymarid_network}

To process the spherical input $\mathcal{S}_0$ and model the relationship on the sphere, we construct a Feature Pyramid Network (FPN) \cite{lin2017feature} with ResNet18\cite{he2016deep} by replacing the conventional 2D convolutions with our newly designed \textit{SPAtial Spherical Convolutions} (SPA-SConvs), giving out the high-level semantic spherical feature map $\mathcal{S} \in \mathbb{R}^{C\times H \times W}$. SPA-SConv is proposed to effectively extract the viewpoint-equivariant spherical features with feature padding and symmetric convolutional operations; we will detail the design of SPA-SConv in Sec. \ref{sec:spatial_spherical_conv}.

For the setting of multiple input signals, \eg, two of $\mathcal{S}_0^{(1)}$ and $\mathcal{S}_0^{(2)}$, we employ multiple spherical ResNets to process them individually, and concatenate their features at each stage of ResNets for feature fusion before feeding them into the top-down pathway in FPN; Fig.~\ref{fig:vi-net} gives the illustration.

\subsection{V-Branch}
\label{subsec:V_branch}

Given the spherical feature map $\mathcal{S}$, a simple solution to obtain the viewpoint rotation $\bm{R}_{vp}$ is first applying a convolution with a big kernel size (\eg, $H \times W$) for a global feature and then directly regressing, as done in \cite{lin2021dualposenet}, which is yet required to handle the global relationship on the whole sphere with huge parameters. To ease this problem, we design the V-Branch to search on the sphere for the canonical zenith direction via binary classification.

Consider the spherical feature map $\mathcal{S} \in \mathbb{R}^{C\times H \times W}$ as $HW$ spherical anchors.
Instead of classifying all of them, we propose to make the task easier by further decoupling it into two sub-tasks, \eg, separate classifications on the azimuthal angle $\varphi$ and the inclination angle $\theta$, 
which also effectively alleviate the severe imbalance on the ratio of positive and negative spherical anchors ($1:HW-1$). Fig. \ref{fig:vi-net} gives the illustration of the V-Branch.

More specifically, we lift the feature channels of $\mathcal{S}$ with a two-layer MLP to yield $\mathcal{S}_{vp} \in \mathbb{R}^{C_{vp}\times H \times W}$. For the learning of azimuthal angle $\varphi$, we max-pool $\mathcal{S}_{vp}$ along the dimension of inclination, giving $\mathcal{F}_{\varphi} \in \mathbb{R}^{C_{vp} \times W}$, which is fed into another MLP to generate the probability map $\mathcal{Y}_{\alpha} \in \mathbb{R}^W$ of $W$ azimuthal anchors. Each element of $\mathcal{Y}_{\alpha}$ indicates the possibility of the anchor being the target. Denoting the index of the element with maximum probability as $w_{max}$, then we have  
\begin{equation}
    \varphi=(w_{max}+0.5) / W \cdot 2\pi.
    \label{eqn:varphi}
\end{equation}
Similarly for the inclination angle $\theta$, we max-pool $\mathcal{S}_{vp}$ along the dimension of azimuth to have $\mathcal{F}_{\theta} \in \mathbb{R}^{C_{vp} \times H}$, and generate the probability map $\mathcal{Y}_{\theta} \in \mathbb{R}^H$ via an MLP. Then $\theta$ could be computed as follows:
\begin{equation}
    \theta= (h_{max}+0.5) / H \cdot  \pi,
    \label{eqn:theta}
\end{equation}
where $h_{max}$ is the index of the element with maximum value in $\mathcal{Y}_{\theta}$. Finally, combining (\ref{eqn:varphi}), (\ref{eqn:theta}) and (\ref{eqn:r_vp}), we obtain the viewpoint rotation $\bm{R}_{vp}=\bm{R}_Z(\varphi) \bm{R}_Y(\theta)$.

\subsection{I-Branch}
\label{subsec:I_branch}

With the viewpoint rotation $\bm{R}_{vp}$, we propose the I-Branch to learn the in-plane rotation $\bm{R}_{ip}$ by viewing the object from the canonical zenith direction to reduce the difficulty of learning, since $\bm{R}_{ip}$ is rather sensitive to $\bm{R}_{vp}$. The first step is to construct a new spherical feature map $\mathcal{S}_{ip} \in \mathbb{R}^{C\times H \times W}$ after rotating the observed system by $\bm{R}_{vp}$ to align the zenith directions of both systems, \eg, aligning the $Z$-axis in $XYZ$ system with the $Z^{\prime}$-axis in $X^{\prime}Y^{\prime}Z^{\prime}$ system (\cf Fig. \ref{fig:intro}). The viewpoint-equivariance of $\mathcal{S}$ makes it possible to realize the transformation in the feature space to obtain $\mathcal{S}_{ip}$.

For a regular spherical map with resolution of $H\times W$, we denote the center points of all $HW$ discrete anchors as a point set $\mathcal{G} = \{\bm{g} \}$. When we rotate the point set $\mathcal{P}=\{\bm{p}\}$ with $\bm{R}_{vp}$ to be $\mathcal{P}^{\prime}=\{\bm{p}^{\prime}\}=\{\bm{R}_{vp}^T\bm{p}\}$, the anchor points of $\mathcal{S}$ are also rotated to be $\mathcal{G}^{\prime} = \{\bm{g}^{\prime}\} = \{\bm{R}_{vp}^T\bm{g} \}$; we denote the feature of $\bm{g}^{\prime}$ as $\mathcal{S}^{\bm{g}^{\prime}}$. To view an object from the canonical zenith direction, a new regular spherical feature map $\mathcal{S}_{ip}$ should be constructed for the transformed $\mathcal{P}^{\prime}$, based on the anchor points $\mathcal{G}$ with canonical orders. For $\mathcal{S}_{ip}$, we generate its feature on each anchor point $\bm{g}$, denoted as $\mathcal{S}_{ip}^{\bm{g}}$, from $\mathcal{S}$ by using the weighted interpolation of point features \cite{qi2017pointnet++} as follows:
\begin{equation}
    \mathcal{S}_{ip}^{\bm{g}} = \frac{\sum_{i=1}^k a_i \mathcal{S}^{\bm{g}_i^{\prime}}}{\sum_{i=1}^k a_i},
    \label{eqn:point_interpolation}
\end{equation}
where  $a_i = \frac{1}{|| \bm{g}- \bm{g}_i^{\prime}||^2}$ is the weight for interpolation measured by point distance, and $\{\bm{g}_i^{\prime}\}_{i=1}^k \subset \mathcal{G}^{\prime}$ is the $k$ nearest points of $\bm{g}$.

After the feature transformation from $\mathcal{S}$ to $\mathcal{S}_{ip}$ via (\ref{eqn:point_interpolation}), we use several stacked strided convolutions to reduce the resolution of $\mathcal{S}_{ip}$ and extract a global feature for the regression of $\bm{R}_{ip}$; Fig. \ref{fig:vi-net} also gives the illustration. The continuous 6D representation of rotation \cite{DBLP:journals/corr/abs-1812-07035} is used as the output of regression, and then transformed to the rotation matrix $\bm{R}_{ip}$. We note that $\bm{R}_{ip}$ here is not restricted to have only one degree-of-freedom (\ie, the angle $\beta$ in (\ref{eqn:r_ip})), and is learned in $SO(3)$ instead as the combination of the residual viewpoint rotation and the exact in-plane rotation, since $\bm{R}_{vp}$ generated by V-Branch is a coarse prediction of the exact viewpoint rotation with discretized azimuthal and inclination angles.

\subsection{Training of VI-Net}
\label{subsec:objective}

For V-Branch, we use focal loss \cite{lin2017focal} on top of two binary classifiers, given the ground truth labels $\hat{\mathcal{Y}}_{\varphi} \in \mathbb{R}^W$ and $\hat{\mathcal{Y}}_{\theta} \in \mathbb{R}^H$, as follows: 
\begin{equation}
    \mathcal{L}_{vp} = \mathcal{D}_{FL}(\mathcal{Y}_{\varphi}, \hat{\mathcal{Y}}_{\varphi}) + \mathcal{D}_{FL}(\mathcal{Y}_{\theta}, \hat{\mathcal{Y}}_{\theta}),
    \label{eqn:loss_vp}
\end{equation}
where
\begin{equation}
    \mathcal{D}_{FL}(\mathcal{Y}, \hat{\mathcal{Y}}) = \frac{1}{M} \sum_{i=1}^{M} -\alpha (1-y_{i,t})^\gamma\text{log}(y_{i,t}),
    \label{eqn:focalloss}
\end{equation}
and
\begin{equation}
    y_{i,t} = \left\{\begin{matrix}
        y_i &  \text{if} &   \hat{y}_i = 1 \\
        1-y_i & \text{if} & \hat{y}_i = 0
      \end{matrix}\right. ,
\end{equation}
with $\mathcal{Y} = \{y_i\}_{i=1}^M$ and $\hat{\mathcal{Y}} = \{\hat{y}_i \in \{0,1\}\}_{i=1}^M$. $\alpha$ is the weighting factor, and $\gamma$ denotes the exponent of the modulating factor.

Given the ground truth rotation $\hat{\bm{R}}$, we supervise the final prediction $\bm{R} = \bm{R}_{vp}\bm{R}_{ip}$ on top of the I-Branch as follows:
\begin{equation}
    \mathcal{L}_{ip} = || \bm{R} - \hat{\bm{R}} || = || \bm{R}_{vp}\bm{R}_{ip} - \hat{\bm{R}} ||.
    \label{eqn:loss_ip}
\end{equation}

Combining (\ref{eqn:loss_vp}) and (\ref{eqn:loss_ip}) gives the following optimization problem to train our VI-Net:
\begin{equation}
    \min \mathcal{L} = \mathcal{L}_{ip} + \lambda \mathcal{L}_{vp},
    \label{eqn:objective}
\end{equation}
where $\lambda$ is the balanced parameter.

\section{Spatial Spherical Convolution}
\label{sec:spatial_spherical_conv}

\begin{figure}[t]
  \centering
   \includegraphics[width=1.0\linewidth]{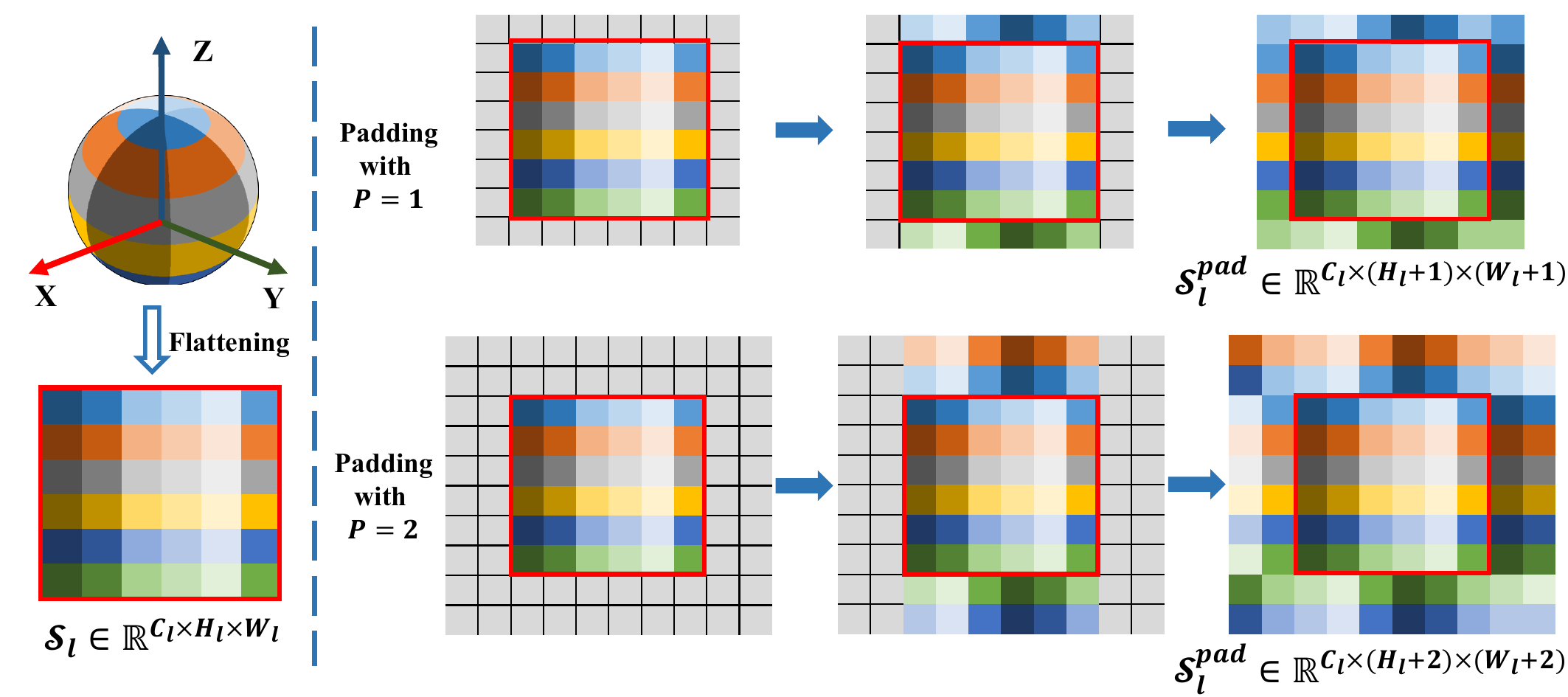}
   \caption{An illustration of the feature padding in SPA-SConv. Best view in the electronic version.}
   \label{fig:spa-sconv}
\end{figure}

We base the feature learning on the sphere for rotation estimation by building the signals as spherical feature maps with regular 2D spatial sizes. Such representations make it possible to process spherical signals via 2D convolutions. However, directly applying 2D convolutions could not realize continuous feature learning on the sphere, resulting in the boundary problem; for example, $S_{0}(h,1)$ has a long distance to $S_{0}(h,W)$ on the feature map $\mathcal{S}_0$ without connection, but their regions border on the sphere actually. Besides, to support the feature transformation in I-Branch for viewing from the zenith direction, convolutions forming the backbone are also required to be viewpoint-equivariant. To settle the above issues, we propose a novel \textit{SPAtial Spherical Convolution}, shortened as \textbf{SPA-SConv}, to extract viewpoint-equivariant features continuously on the sphere, which could be flexibly adapted into existing convolutional network architectures, \eg, FPN in Sec. \ref{subsec:spherical_feature_prymarid_network}.

Given the input spherical feature map $\mathcal{S}_{l} \in \mathbb{R}^{C_l\times H_l \times W_l}$ and the parameters \wrt convolution (\eg, the kernel size $K$, stride $s$, the number of output channels $C_{l+1}$), we implement our SPA-SConv in two steps, including 1) padding $\mathcal{S}_{l}$ to $\mathcal{S}_{l}^{pad}\in \mathbb{R}^{C_l\times (H_l+2P) \times (W_l+2P)}$ with $P=(K-1)/2$, and 2) applying symmetric convolutional operations to $\mathcal{S}_{l}^{pad}$, based on conventional 2D convolution with no padding, to generate the output spherical feature map $\mathcal{S}_{l+1} \in \mathbb{R}^{C_{l+1}\times H_{l+1} \times W_{l+1}}$, where $H_{l+1}=\lfloor H_{l}/s \rfloor$ and $W_{l+1}=\lfloor W_{l}/s \rfloor$. \footnote{For simplicity, We assume the length and width are both $K$ ($K$ is an odd number) for the 2D convolutional kernel. }

Fig. \ref{fig:spa-sconv} illustrates the padding of $\mathcal{S}_{l}$ to $\mathcal{S}_{l}^{pad}$. Firstly, we have $\mathcal{S}_{l}$ at the center of $\mathcal{S}_{l}^{pad}$:
\begin{equation}
    \mathcal{S}_{l}^{pad}(h+P, w+P) = \mathcal{S}_{l}(h, w),
\end{equation}
for $\forall h=1,2,\dots,H_{l}$ and $\forall w=1,2,\dots,W_{l}$. Next, we pad $\mathcal{S}_{l}$ along inclination ($\theta$-axis in the sphere coordinate system) as follows: 
\begin{equation}
    \begin{matrix}
    \mathcal{S}_{l}^{pad}(p, w+P)  = \mathcal{S}_{l}^{pad}(2P-p+1, w^{\prime}), \\
     \text{and}  \mathcal{S}_{l}^{pad}(H_l+P+p, w+P)  = \mathcal{S}_{l}^{pad}(H_l+P-p+1, w^{\prime}),
    \end{matrix}
\end{equation}
where 
\begin{equation}
    w^{\prime} = \left\{\begin{matrix}
        w + W_l/2 +P &  \text{if}  &  w \le W_l/2 \\
        w - W_l/2 +P & & \text{otherwise}
      \end{matrix}\right. ,
\end{equation}
for $\forall p=1,2,\dots,P$ and $\forall w=1,2,\dots,W_{l}$. Finally we pad the feature map along azimuth ($\varphi$-axis in the sphere coordinate system): 
\begin{equation}
    \begin{matrix}
    & \mathcal{S}_{l}^{pad}(h, p) = \mathcal{S}_{l}^{pad}(h, W_l+p),\\
    \text{and}  & \mathcal{S}_{l}^{pad}(h, W_l+P+p) = \mathcal{S}_{l}^{pad}(h, P+p),
    \end{matrix}
\end{equation}
for $\forall p=1,2,\dots,P$ and $\forall h=1,2,\dots,H_{l}+2P$. 

With the padded $\mathcal{S}_{l}^{pad}$, we could thus exploit the 2D convolution to realize the symmetric convolutional operations for the extraction of viewpoint-equivariant features as follows:
\begin{equation}
    \mathcal{S}_{l+1} = \texttt{Max} (\texttt{Conv}(\mathcal{S}_{l}^{pad}; \kappa_l), \texttt{Conv}(\mathcal{S}_{l}^{pad}; \texttt{Flip}(\kappa_l))),
\end{equation}
where \texttt{Conv}, \texttt{Flip}, and \texttt{Max} denotes the 2D convolutional operation, horizontal flip, and element-wise max-pooling, respectively, with $\kappa_l$ denoting the weight of the convolution. \texttt{Max} serves as the symmetric function to aggregate the features \cite{qi2017pointnet} and keeps the property of viewpoint-equivariance; we provide the proof of this property in the supplemental material, where the pseudo code is also included to help understand the implementation of SPA-SConv.

\section{Category-level 6D Object Pose Estimation}
\label{sec:task}

The task of category-level 6D object pose estimation is to estimate the poses of unknown objects \wrt categorical canonical spaces, including the learning of rotation $\bm{R} \in SO(3)$, translation $\bm{t}\in \mathbb{R}^3$ and size $\bm{s}\in \mathbb{R}^3$ for each object. CAD models are not available during testing time. To confirm the efficacy of our VI-Net on rotation estimation, we apply VI-Net to this challenging task and realize the target with three stages. Specifically, given an RGB-D image of a cluttered scene, we firstly employ MaskRCNN \cite{MasRCNN} to segment the objects of interest out; for each cropped object with a point set $\mathcal{P} = \{\bm{p}\}$, we then estimate its translation $\bm{t}$ and size $\bm{s}$ via a simple and lightweight network of PointNet++ (PN2) \cite{qi2017pointnet++} with multi-scale grouping; in the final stage, we normalize the point set as $\mathcal{P^\prime} = \{(\bm{p}-\bm{t})/||\bm{s}||\}$ and feed it into our VI-Net for the estimation of rotation $\bm{R}$.

\subsection{Experimental Setups}

\vspace{0.1cm}
\noindent \textbf{Datasets} Experiments are conducted on the benchmarking CAMERA25 and REAL275 datasets \cite{NOCS}, both of which enjoy the same object categories. CAMERA25 is a synthetic dataset generated by a context-aware mixed reality approach; it contains 300,000 RGB-D images of 1,085 objects in total, with 25,000 images of 184 objects set aside for validation. REAL275 is a challenging real-world dataset with various lighting conditions and occlusions; for this dataset, 4,300 images of 7 scenes are split as a training set, while the other 2,750 images of 6 scenes form the test set. Following \cite{NOCS}, we mix both training sets to optimize the parameters of VI-Net.

\vspace{0.1cm}
\noindent \textbf{Implementation Details} We use the instance masks generated by MaskRCNN \cite{MasRCNN} and provided by \cite{SPD} to segment objects of interest out. For data processing, we use point-wise radial distances and RGB values as two input signals. The input resolution of $\mathcal{S}_0$ is set as $H_0\times W_0=64\times 64$, while that of $\mathcal{S}$ is $H\times W=32\times 32$. The point number of the sampled point set is $N=2,048$, and $k$ in (\ref{eqn:point_interpolation}) is set as $3$. For the training of VI-Net, we set $\alpha=0.5$ and $\gamma=2.0$ in (\ref{eqn:focalloss}); the balanced parameter $\lambda$ in (\ref{eqn:objective}) is set as $100$. We use ADAM to train VI-Net for a total of $200,000$ iterations. The learning rate is initialized as $0.001$, with a cosine annealing schedule used afterward, while the data size of a min-batch is set as 128.  

\vspace{0.1cm}
\noindent \textbf{Evaluation Metrics} We report the mean Average Precision (mAP) of $n\degree m$ cm for 6D pose estimation, which denotes the pose precision with rotation error less than $n\degree$ and translation error less than $m$ cm. Following \cite{NOCS,liu2022catre}, we also report mAP of Intersection over Union (IoU$_x$) with a threshold of $x\%$ for object detection.

\begin{table*}
  \centering
  \resizebox{\textwidth}{!}{
  \begin{tabular}{l|c|ccccc|ccccc}
    \toprule
    \multirow{2}{*}{Method}  & Use of  & \multicolumn{5}{c|}{REAL275} & \multicolumn{5}{c}{CAMERA25} \\
    \cline{3-12}
    & Shape Priors & IoU$_{75} * $ &  5\degree 2cm & 5\degree 5cm & 10\degree 2cm & 10\degree 5cm  &  IoU$_{75} * $  & 5\degree 2cm & 5\degree 5cm & 10\degree 2cm & 10\degree 5cm\\
    \midrule
    \midrule
    NOCS \cite{NOCS}   & $\times$ & 9.4 & 7.2 & 10.0 & 13.8 & 25.2 & 37.0 & 32.3 & 40.9 & 48.2 & 64.6 \\
    FS-Net \cite{FSNet}   & $\times$ & -  & - & 28.2 & - & 60.8 & - & - & - & -  & -  \\
    DualPoseNet \cite{lin2021dualposenet} & $\times$  & 30.8  & 29.3 & 35.9 & 50.0 & 66.8 & 71.7 &  64.7 & 70.7  & 77.2 & 84.7\\
    GPV-Pose \cite{GPV}  &$\times$  & -   & 32.0 & 42.9 & - & 73.3  & -   & 72.1 & 79.1 & - &  89.0 \\ 
    SS-ConvNet \cite{SS-Conv} & $\times$ & - & 36.6 & 43.4  & 52.6 & 63.5 & - & - & -& - & -\\
    \hline
    \textbf{PN2 + VI-Net (Ours)}  & $\times$  & \textbf{48.3} & \textbf{50.0} & \textbf{57.6} & \textbf{70.8} & \textbf{82.1}  & \textbf{79.1} & \underline{74.1} & \textbf{81.4} & 79.3 & 87.3  \\
    \midrule
    \midrule
    SPD \cite{SPD}   & \checkmark   & 27.0 & 19.3 & 21.4 & 43.2 & 54.1  & 46.9 &  54.3 & 59.0 & 73.3  & 81.5  \\ 
    CR-Net \cite{CRNet}   & \checkmark   & 33.2 & 27.8 & 34.3 & 47.2 & 60.8 & 75.0 & 72.0 & 76.4  & 81.0 &  87.7  \\
    CenterSnap-R \cite{irshad2022centersnap} &  \checkmark & - & - & 29.1 & - & 64.3 & - & - & 66.2 & - & 81.3 \\
    ACR-Pose \cite{fan2021acr}  & \checkmark  & - & 31.6 & 36.9 & 54.8 & 65.9 & - & 70.4 & 74.1 & 82.6 & 87.8  \\
    SAR-Net \cite{lin2022sar}  & \checkmark  & - & 31.6 & 42.3 & 50.3 & 68.3  & - & 66.7 &  70.9 & 75.3 &  80.3\\
    SSP-Pose \cite{zhang2022ssp}  & \checkmark   & - & 34.7 & 44.6 & - & 77.8 & - & 64.7 & 75.5 & - &  87.4 \\
    SGPA \cite{SGPA}   & \checkmark   & 37.1 & 35.9 & 39.6 & 61.3 & 70.7 & 69.1 &  70.7 & 74.5 & \underline{82.7} & 88.4 \\
    RBP-Pose \cite{zhang2022rbp}  & \checkmark  & - &  38.2 & 48.1 & 63.1 & \underline{79.2}  & - & 73.5 &79.6 & 82.1   & \textbf{89.5}\\
    SPD + CATRE \cite{liu2022catre} & \checkmark & \underline{43.6} & 45.8 & \underline{54.4} & 61.4 & 73.1  & \underline{76.1}  & \textbf{75.4} & \underline{80.3} & \textbf{83.3} & \underline{89.3} \\
    DPDN \cite{DPDN}  & \checkmark   & - & \underline{46.0} & 50.7 & \underline{70.4} & 78.4 & - & - & -& - & -\\
    \bottomrule
  \end{tabular}}
  \vspace{0.1cm}
  \caption{Quantitative comparisons of different methods for category-level 6D object pose estimation on REAL275 and CAMERA25 datasets \cite{NOCS}. Overall best results are in \textbf{bold} and the second best results are \underline{underlined}. `*' denotes the IoU metrics used in \cite{liu2022catre} rather than those in \cite{NOCS}.}
  \label{tab:real275_sota}
\end{table*}

\begin{figure*}[h]
  \centering
   \includegraphics[width=0.98\linewidth]{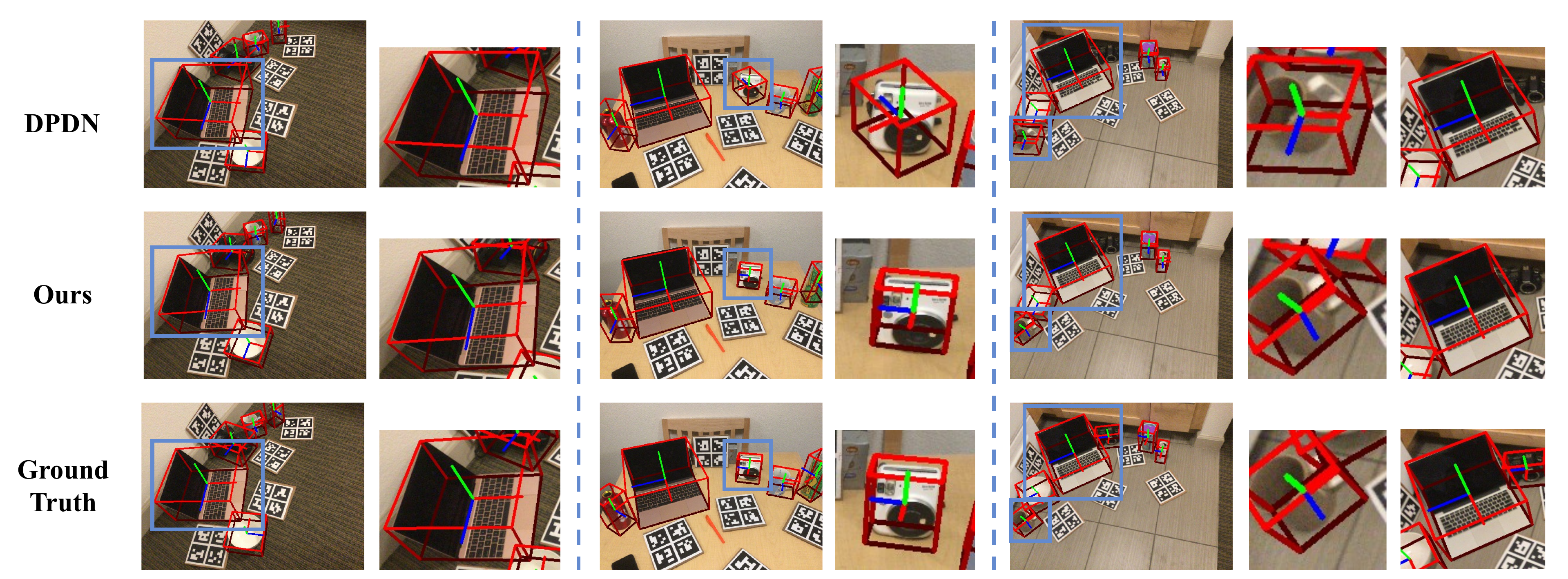}
   \caption{Qualitative comparisons between the state-of-the-art method of DPDN \cite{DPDN} and our proposed one on REAL275 dataset \cite{NOCS}.}
   \label{fig:vis-sota}
\end{figure*}

\begin{figure}[h]
  \centering
   \includegraphics[width=0.98\linewidth]{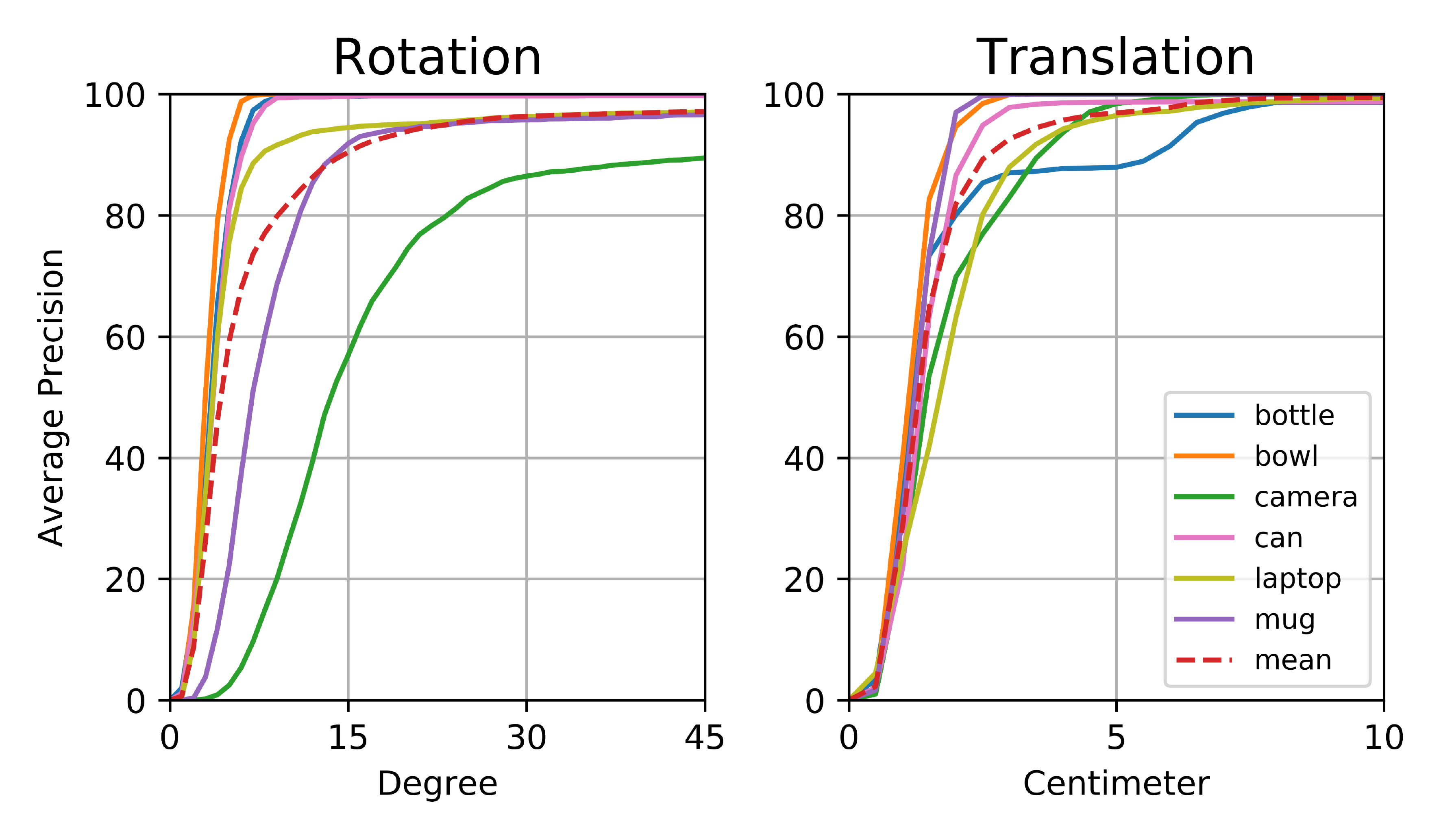}
   \caption{Plottings of per-category average precision versus different rotation/translation error thresholds for our proposed method on REAL275 dataset \cite{NOCS}.}
   \label{fig:per-cat}
\end{figure}

\begin{table}
  \centering
  \resizebox{0.47\textwidth}{!}{
  \begin{tabular}{l|c|cc}
    \toprule
     & R.D.  & 5\degree 2cm & 5\degree 5cm  \\
    \midrule
    Baseline1: Avg Pool + MLP & $\times$ & 45.0 & 51.8  \\
    Baseline2: Flattening + MLP & $\times$& 42.7 & 48.8  \\
    VI-Net & \checkmark  & \textbf{50.0} & \textbf{57.6} \\
    \bottomrule
  \end{tabular}}
  \vspace{0.1cm}
  \caption{Ablation studies of the variants of our VI-Net with or without rotation decomposition (R.D.) on REAL275 dataset \cite{NOCS}.}
  \label{tab:ab-decom}
\end{table}

\begin{table*}
  \centering
  \resizebox{0.85\textwidth}{!}{
  \begin{tabular}{c|c|cccc}
    \toprule
    Variants of V-Branch & Feature Trans. in I-Branch    & 5\degree 2cm & 5\degree 5cm & 10\degree 2cm & 10\degree 5cm\\
    \midrule
    Direct Regression  & $\times$  & 26.3 & 28.9 & 54.1 & 61.1   \\
    Binary Classification (1-branch) & $\times$   & 30.8 & 34.4 & 65.5 & 75.3 \\
    Binary Classification (2-branch) & $\times$  & 33.7 & 37.9 & 65.4 & 75.4 \\
    \hline
    Direct Regression  & \checkmark   & 43.1 & 48.3 & 68.0 & 78.1 \\
    Binary Classification (1-branch) & \checkmark  & 49.3 & 56.9 & 69.4 & 79.7  \\
    Binary Classification (2-branch) & \checkmark   & \textbf{50.0} & \textbf{57.6} & \textbf{70.8} & \textbf{82.1}\\
    \bottomrule
  \end{tabular}}
  \vspace{0.1cm}
  \caption{Ablation studies of the variants of the V-Branch and I-Branch in our VI-Net on REAL275 dataset \cite{NOCS}.}
  \label{tab:ab-vi}
\end{table*}

\begin{figure*}[h]
  \centering
   \includegraphics[width=1.0\linewidth]{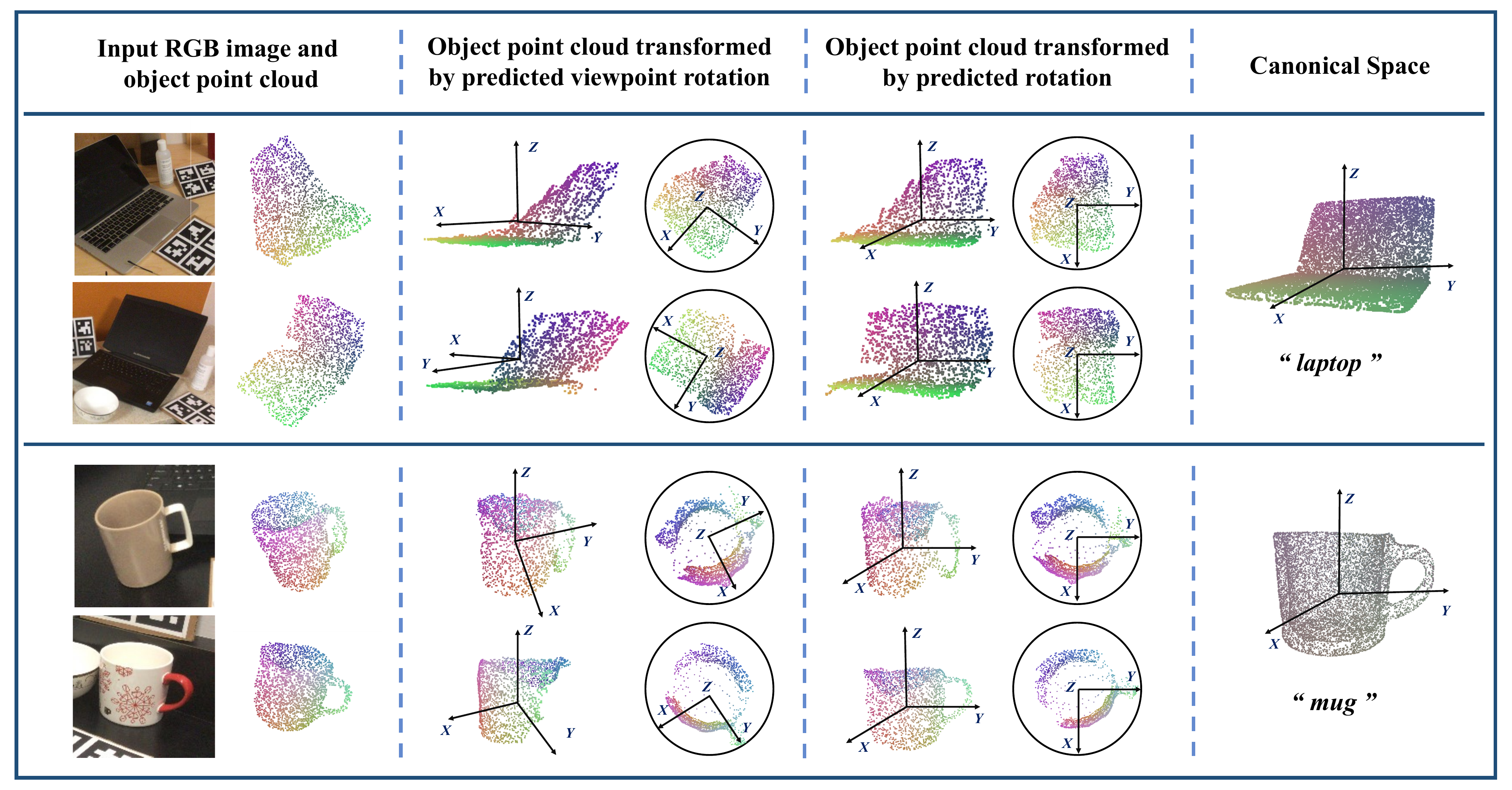}
   \caption{Qualitative results of the predicted viewpoint rotation $\bm{R}_{vp}$ from V-Branch and the final output rotation $\bm{R}$ from VI-Net.}
   \label{fig:vis-viewpoint}
\end{figure*}

\begin{table}
  \centering
  \resizebox{0.47\textwidth}{!}{
  \begin{tabular}{c|c|c|cc}
    \toprule
    Type of & \multirow{2}{*}{Padding} & Symmetric & \multirow{2}{*}{5\degree 2cm} & \multirow{2}{*}{5\degree 5cm}\\
    Convolution &    & Operation &  & \\
    \midrule
    SPE-SConv \cite{esteves2018learning} & $-$ & $-$ &  35.1 & 40.8\\
    \hline
    \multirow{3}{*}{SPA-SConv} & $\times$ & $\times$   & 46.9 &  53.2\\
    & $\times$ & \checkmark & 48.5 & 55.5  \\
    & \checkmark & \checkmark  & \textbf{50.0} & \textbf{57.6} \\
    \bottomrule
  \end{tabular}}
  \vspace{0.1cm}
  \caption{Quantitative comparisons of SPE-SConv \cite{esteves2018learning} and variants of our proposed SPA-SConv on REAL275 dataset \cite{NOCS}.}
  \label{tab:ab-conv}
\end{table}

\begin{table}
  \centering
  \begin{tabular}{l|cccc}
    \toprule
     &  5\degree 2cm & 5\degree 5cm & 10\degree 2cm & 10\degree 5cm\\
    \hline
    VI-Net$_{+ts}$ & 45.0 & 56.0 & 65.9 & 80.5 \\
    PN2 + VI-Net  & \textbf{50.0} & \textbf{57.6} & \textbf{70.8} & \textbf{82.1}\\
    \bottomrule
  \end{tabular}
  \vspace{0.1cm}
  \caption{Quantitative comparisons between VI-Net$_{+ts}$ and PN2 for translation and size estimation on REAL275 dataset \cite{NOCS}.}
  \label{tab:ab-ts}
\end{table}

\begin{table}
  \centering
  \begin{tabular}{l|cccc}
    \toprule
    Data Type &  5\degree 2cm & 5\degree 5cm & 10\degree 2cm & 10\degree 5cm\\
    \hline
    Depth & 43.0 & 52.1 & 64.3 & 77.0 \\
    RGB $+$ Depth  & \textbf{50.0} & \textbf{57.6} & \textbf{70.8} & \textbf{82.1}\\
    \bottomrule
  \end{tabular}
  \vspace{0.1cm}
  \caption{Quantitative comparisons of different input data types of VI-Net on REAL275 dataset \cite{NOCS}.}
  \vspace{-0.2cm}
  \label{tab:ab-rgbd}
\end{table}

\subsection{Comparisons with Existing Methods}

To verify the efficacy of VI-Net, we conduct experiments on REAL275 and CAMERA25 datasets \cite{NOCS} to compare with the existing methods, including those without the use of shape priors \cite{NOCS,FSNet, lin2021dualposenet, GPV,SS-Conv}, and those using shape priors \cite{SPD,CRNet,fan2021acr,lin2022sar,zhang2022ssp,SGPA,zhang2022rbp,liu2022catre}. We firstly employ the lightweight network of PointNet++ (PN2) \cite{qi2017pointnet++} for translation and size estimation, before using VI-Net for rotation estimation, and shape priors (as well as CAD object models) are not used in our method. Quantitative results of different methods are given out in Table \ref{tab:real275_sota}.

In terms of pose estimation, our method outperforms those without the use of shape priors on all the evaluation metrics of $n \degree m$ cm with large margins on REAL275 dataset, as shown in the table. For example, on the metric of $5\degree 2$ cm, our method refreshes the result to reach the mAP of $50.0\%$, which exceeds the $SE(3)$-equivariant SS-ConvNet \cite{SS-Conv} and the DualPoseNet \cite{lin2021dualposenet} based on $SO(3)$-equivariant SPE-SConv \cite{esteves2018learning} by $13.4\%$ and $20.7\%$, respectively. We note that DualPoseNet also learns object poses on the sphere by applying convolutions with large kernels to model the global relationship, while our VI-Net exploits Spherical FPN with SPA-SConv to model local relationships along hierarchy more efficiently, and achieves much better performance with the individual learning of the factorized rotations. Compared to the methods using shape priors, VI-Net also possesses powerful competition on REAL275 dataset; taking DPDN \cite{DPDN} as example, our method improves the results by $4.0\%$ mAP on $5\degree 2$cm and $6.9\%$ mAP on $5\degree 5$cm, verified by the quantitative comparisons in Fig. \ref{fig:vis-sota}. We provide per-category comparisons of our method in terms of rotation error and translation error on REAL275 dataset in Fig. \ref{fig:per-cat}. On CAMERA25 dataset, as shown in Table \ref{tab:real275_sota}, our method also achieves comparable results, especially on the more strict metrics, \eg, IoU$_{75}$. 


\subsection{Ablation Studies and Analyses}

To confirm the efficacy of our designs, we conduct experiments to compare the variants of VI-Net in different aspects on REAL275 dataset \cite{NOCS}.

\vspace{0.15cm}
\noindent \textbf{Efficacy of Rotation Decomposition in VI-Net} The foundation of VI-Net is the decomposition of rotation into a viewpoint rotation and an in-plane rotation, which guides to factorize the task into two sub ones to reduce the learning difficulty in the whole $SO(3)$ space. We thus firstly verify the mechanism of rotation decomposition in VI-Net by building two baselines, which regress the rotation $\bm{R}$ directly from the global features learned by either globally average-pooling or flattening the spherical feature map $\mathcal{S}$. Both baselines employ the same backbone of Spherical FPN as VI-Net, and their results are listed in Table \ref{tab:ab-decom}, where VI-Net outperforms both of them with large margins on all the evaluation metrics, confirming the effectiveness of rotation decomposition in pose estimation.

\vspace{0.15cm}
\noindent \textbf{Efficacy of Separate Binary Classifications in V-Branch} We compare three variants of V-Branch, including i) direct regression of $\bm{R}_{vp}$ from the globally average-pooled feature from $\mathcal{S}$, binary classification on the whole spherical map, and the two separate binary classifications we introduced in Sec. \ref{subsec:V_branch}. As shown in Table \ref{tab:ab-vi}, among three variants, the last two ones make more precise predictions by searching on the sphere for the canonical zenith direction via binary classification, which simplifies the regression task as the recognition one(s). The last one further improves the pose precision via the decoupled recognition of the azimuthal angle $\varphi$ and the inclination angle $\theta$, since binary classification on the whole spherical map faces the severe imbalance on the ratio of positive and negative samples (1:1023 in our experiments), even though we use the focal loss (\ref{eqn:focalloss}) to alleviate this problem. we also note that, comparing with the results in Table \ref{tab:ab-decom}, the key to the success of rotation decomposition for VI-Net lies in exactly the binary classification of V-Branch, \ie, the correlation on the rotation decomposition and sphere.  Finally, We visualize some examples in Fig. \ref{fig:vis-viewpoint}, where object point clouds transformed by the predicted viewpoint rotations from V-Branch are presented. 

\vspace{0.15cm}
\noindent \textbf{Efficacy of Feature Transformation in I-Branch} In Table \ref{tab:ab-vi}, we also verify, with three different variants of V-Branch, the advantages of feature transformation in I-Branch, which transforms in feature space with viewpoint rotation $\bm{R}_{vp}$ to make the network ``see" the object from the canonical zenith direction, and thus reduces the difficulty in estimating the in-plane rotation $\bm{R}_{ip}$. Applying the feature transformation also makes I-Branch aware of the learned $\bm{R}_{vp}$, and could include the learning of the residual viewpoint rotation in I-Branch to refine predictions from V-Branch.

\vspace{0.15cm}
\noindent \textbf{Efficacy of Spatial Spherical Convolutions (SPA-SConv)} We propose SPA-SConv to realize continuous convolution on the sphere, which settles the boundary problems of spherical representations by simply feature padding and realizes the properties of viewpoint-equivariance by the symmetric convolutional operations. In Table \ref{tab:ab-conv}, we demonstrate the advantages of these novel designs individually. As introduced in Sec. \ref{subsec:related_ssconv}, the $SO(3)-$equivariant spherical feature maps learnt by SPE-SConv \cite{esteves2018learning} are less discriminative for the binary classification of V-Branch, even resulting in gradient exploding. Therefore, to avoid gradient exploding, we add 4 additional SPA-Convs in V-Branch for the VI-Net based on SPE-SConvs, and the training thus becomes stable; however, the results based on SPE-SConvs are still less satisfactory compared to those of SPA-SConvs, as shown in Table \ref{tab:ab-conv}, indicating the importance of SPA-SConv in the design of VI-Net.

\vspace{0.15cm}
\noindent \textbf{Effects of Including Translation and Size Estimation in VI-Net} We focus on learning decoupled rotations on spherical inputs converted from centralized point sets, and separate translation and size estimation via PointNet++ (PN2). Another choice (dubbed VI-Net$_{+ts}$) is to use the mean coordinate of the point set as initial translation and include the residual translation/size estimation with an MLP from the spherical features $\mathcal{S}$. As shown in Table \ref{tab:ab-ts}, the performance gap between VI-Net$_{+ts}$ and PN2 is acceptable, motivating us to explore better combined learning in the future.

\vspace{0.15cm}
\noindent \textbf{Effects of Different Input Data Types} The primary input data of VI-Net are depth maps (or point clouds), which are used for computing the point locations on the sphere and further transformed as radial distances to be the input signals. Additionally, other point attributes, \eg, RGB values, could also be the inputs. In Table \ref{tab:ab-rgbd}, we demonstrate that VI-Net effectively learns both geometric and appearance information from Depth maps and RGB images, respectively,  which indicates VI-Net could be used for handling different input types to jointly make the predictions.

\section{Conclusion}

In this paper, we propose a new method of VI-Net for rotation estimation, which bases the feature learning on the sphere, and factorizes the task into two sub-ones with individual heads of V-Branch and I-Branch. V-Branch learns the viewpoint rotation via two separate binary classifications on top of the spherical feature maps, while I-Branch transforms the feature maps with the viewpoint rotation to view objects from the canonical zenith direction for the estimate of in-plane rotation. To support the learning of these two branches, we also propose a novel design of Spatial Spherical Convolution for the extraction of viewpoint-equivariant spherical features. Finally, we apply our VI-Net to the challenging task of category-level 6D object pose estimation and experimental results confirm its efficacy in the regime of high precision of rotation.

\vspace{0.3cm}
\noindent\textbf{Acknowledgements} This work is supported in part by Guangdong R\&D key project of China (No.: 2019B010155001), and the Program for Guangdong Introducing Innovative and Enterpreneurial Teams (No.: 2017ZT07X183).
{\small
\bibliographystyle{ieee_fullname}
\bibliography{egbib}
}

\end{document}